\documentclass{article}

\PassOptionsToPackage{numbers, compress}{natbib}

\usepackage{comment}
\usepackage{placeins}

 \usepackage[dblblindworkshop, final]{neurips_2026}

\usepackage[utf8]{inputenc} 
\usepackage[T1]{fontenc}    
\usepackage{hyperref}       
\usepackage{url}            
\usepackage{booktabs}       
\usepackage{amsfonts}       
\usepackage{nicefrac}       
\usepackage{microtype}      
\usepackage[table]{xcolor}         
\usepackage{amsmath}
\usepackage{amssymb} 
\usepackage{multirow}
\usepackage{booktabs}
\usepackage{tabularx}
\usepackage{graphicx}
\usepackage{todonotes}
\usepackage[pdftex,dvipsnames]{xcolor}
\usepackage[noabbrev, nameinlink]{cleveref}
\usepackage{subcaption}
\usepackage{svg}

\title{Holding a Steady Course: Knowledge Evolution from Expertise and Lessons for Continual Agent Learning}

\author{%
Wenzhe Fan$^{1\,*\,\dagger}$
\And Benjamin Steenhoek$^{2}$
\And Gaurav Mittal$^{2}$
\And Yu Hu$^{2}$
\AND
{\normalfont\small $^{1}$University of Illinois Chicago, IL, USA}
\And
{\normalfont\small $^{2}$Microsoft, Redmond, WA, USA}
}

\workshoptitle{Towards Test-Time Continual Learning Agents}

\begin{document}

\maketitle

\begingroup
\renewcommand{\thefootnote}{*}
\footnotetext{Work done during an internship with Microsoft Code\textbar{}AI.}
\renewcommand{\thefootnote}{\ensuremath{\dagger}}
\footnotetext{Corresponding author: \texttt{wfan23@uic.edu}}
\endgroup

\begin{abstract}
Autonomous agents are increasingly deployed across sequences of related tasks, yet they often fail to retain and improve their knowledge over time. Existing approaches convert interaction trajectories into memories or reusable skills, but experience is not automatically reliable knowledge. Automatically collected skills and memories can propagate errors, mislead, and corrupt useful knowledge. To address this gap, we introduce Knowledge Evolution from Expertise and Lessons (KEEL), an agent-agnostic framework that uses verifier evidence to transform successful behavior into reusable skills and failures into corrective lessons. As new evidence accumulates, individual skill claims can be supported, contradicted, replaced, or retired. On SkillFlow benchmark, KEEL improves task success by 4.2 to 21.7 percentage points over the no-knowledge baseline, compared with gains of 2.4 to 9.6 points from SkillFlow’s default method. Ablations show that skills and lessons are complementary. 
These results show that reliable continual agent learning depends not simply on accumulating experience, but on continually validating and evolving the knowledge derived from it.
Our code is available at: \url{https://github.com/microsoft/KEEL-TTCL26}.
\end{abstract}

\section{Introduction}
Large language model agents increasingly perform complex tasks across diverse
domains \citep{johnston2026shiftagenticaievidence,
baumann2026swechatcodingagentinteractions}.
Yet most agents are commonly developed and evaluated in static episodic settings due to the challenges of enacting realistic multi-turn/multi-session evaluations~\citep{zheng2026lifelong}. As a result, when applied in a \textit{lifelong learning} scenario, where an agent is repeatedly applied for different tasks in a given project or group of projects, agents commonly fail to persist knowledge between tasks~\citep{suzgun2026dynamic,zhang2026skillflow}.
A growing body of work enables agents to improve from experience without updating the parameters of their underlying language models. These include methods for distilling interactions into reusable natural-language text~\citep{shinn2023reflexion,zhao2024expel}, creating executable and reusable skill libraries \citep{wang2023voyager,pmlr-v267-wang25bx,zheng2025skillweaver}, and combining the two approaches to maintain complementary representation of individual experiences and higher-level skills~\citep{suzgun2026dynamic,zhang2026agentic, zhou2026memento,jiang2026xskill,lin2026muse}.
Recent methods
further improve skill reliability through additional validation executions
\citep{shang2026hypothesis}, or combine repeated skill evaluation with policy
optimization \citep{mao2026learning,shen2026dynamic}.

While these studies demonstrate that persisted knowledge can support improvement across interactions, they suffer from a critical limitation: \textbf{experience is not automatically reliable knowledge}. In particular, a
successful task outcome does not guarantee that the underlying trajectory
contains reliable behavior
\citep{sahoo2026agentlens}. An interaction trajectory may interleave useful behavior with ineffective or harmful actions, and may satisfy some verifier checks while violating others. 
A single task-level outcome is too coarse to determine which parts of a trajectory deserve consolidation.
Persistent memory can propagate errors~\citep{xiong2026memory}, induce negative transfer or compromise retention~\citep{hu2026continual}, and corrupt useful knowledge even when the underlying experiences are correct~\citep{zhang2026useful}.
Auto-generated skills can introduce implementation faults or override user instructions~\citep{dong2026agentskillsharmfulempirical}, and brittle assumptions can make them unsuitable for future tasks~\citep{jiang2026demystifyingagentskillsworkuntil}.
Together, these findings identify a key challenge: agents must incorporate new knowledge \textit{without corrupting their existing knowledge base}.
This involves determining which evidence warrants promoting behavior into persisted knowledge, 
which observations should be retained as lessons, and which individual knowledge claims should be supported, revised, or retired as new evidence accumulates.

To address these challenges, we propose \textbf{K}nowledge \textbf{E}volution from \textbf{E}xpertise and \textbf{L}essons (KEEL), a non-parametric agent continual-learning framework 
for maintaining procedural and corrective knowledge grounded in verifiable task experience. 
KEEL consolidates sequential interaction outcomes into two complementary forms of knowledge: \textit{expertise} and \textit{lessons}. Expertise captures successful behavior as reusable procedural skills, while lessons record mistakes, violated constraints, and potential recovery strategies.
Beyond this skill/lesson decomposition, KEEL introduces fine-grained, evidence-grounded knowledge management that connects verifier outcomes to specific trajectory behaviors and skill claims. 
Rather than treating skills as indivisible knowledge units, KEEL maintains an explicit claim-level lifecycle that supports evidence-based promotion, contradiction, replacement, and retirement while preserving provenance.

We evaluate KEEL using Terminus-2 on SkillFlow \citep{zhang2026skillflow}, a benchmark containing 166 tasks organized into 20 ordered workflow families. Within each family, tasks are presented sequentially in increasing order of difficulty, and the agent carries knowledge acquired from earlier tasks into later ones. This setting evaluates whether an agent can effectively learn from experience and transfer that knowledge over time.
Across all three complete model comparisons, KEEL improves pass rates over No Skills by 4.2 to 21.7 percentage points, compared with improvements of 2.4 to 9.6 percentage points achieved by SkillFlow’s default method over the same baseline.
With Claude Opus 4.6, KEEL increases the pass rate from 44.0\% with No Skills
to 65.7\%, compared with 53.6\% for SkillFlow's default skill-evolution method.
Ablations show that skills and lessons provide complementary benefits. We further analyze the
associated inference cost and the effect of the underlying model and reasoning configuration. 
Experimental results demonstrate the effectiveness of KEEL's evidence-grounded knowledge-management framework for continual agent learning.

Our contributions are threefold:
\begin{itemize}
    \item We introduce an evidence-grounded framework for continual agent learning that extracts reusable skills and lessons from agent experience while keeping model parameters fixed.

    \item We develop a claim-level lifecycle that uses verifier and trajectory
    evidence to promote, contradict, replace, and retire knowledge while
    preserving its provenance.

    \item We evaluate KEEL across multiple models and reasoning configurations, finding consistent improvements over baselines. We also analyze inference costs and the contributions of individual components.
\end{itemize}

\section{Method}
\label{sec:method}

\subsection{Problem Formulation}
\label{sec:problem-setting}

Let \(\mathcal{F}_i=(P_{i,1},\ldots,P_{i,n_i})\) denote an ordered workflow
family. After task \(P_{i,j}\), KEEL maintains the family-specific knowledge
state
\[
\mathcal{K}_i^{(j)}
=
\left(\mathcal{S}_i^{(j)},\mathcal{M}_i^{(j)}\right),
\]
where \(\mathcal{S}_i^{(j)}\) contains reusable skills and
\(\mathcal{M}_i^{(j)}\) contains diagnostic and corrective lessons.
Initially,
\(\mathcal{K}_i^{(0)}=(\varnothing,\varnothing)\). 
For task \(P_{i,j}\), an Actor with fixed model parameters receives a
task-conditioned view of \(\mathcal{K}_i^{(j-1)}\).
KEEL updates this knowledge using the trajectory and verifier feedback, making the updates available only to subsequent tasks in the same family.

\begin{figure}[!t]
\centering
\includegraphics[width=1.0\textwidth]{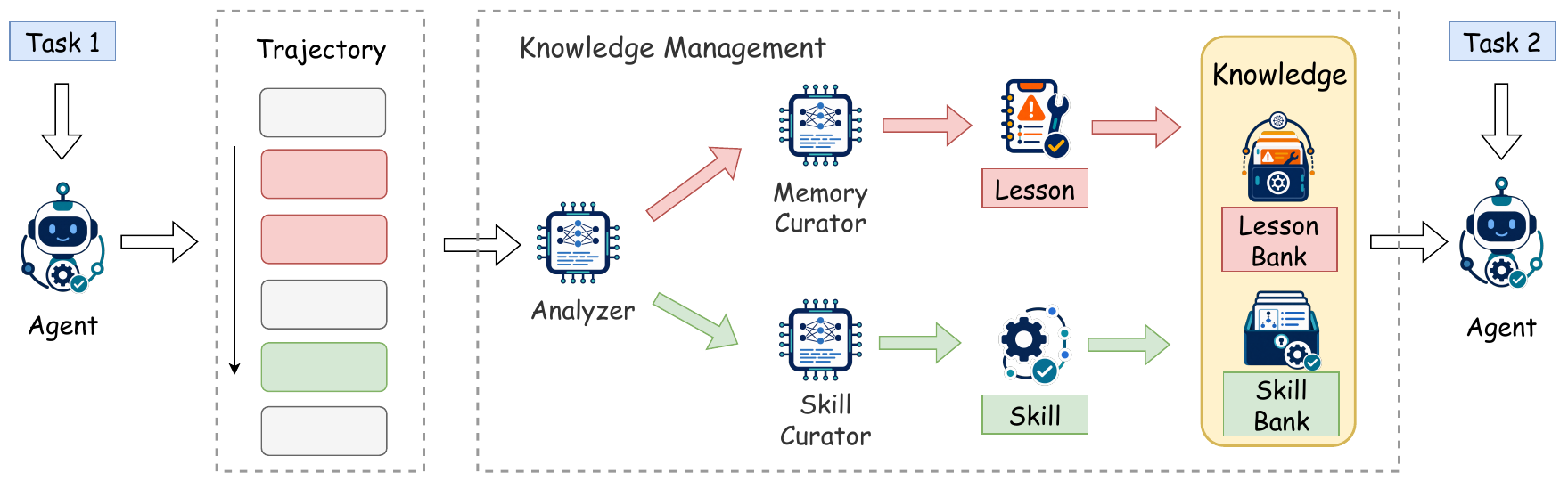}
\caption{Overview of KEEL. After completing each task, the agent's trajectory is analyzed to extract reusable knowledge from its experience. The Memory Curator distills corrective lessons from observed mistakes and failures, while the Skill Curator extracts procedural skills from successful behaviors. These lessons and skills are made available to the agent and continuously accumulate and evolve across subsequent tasks, enabling continual improvement from experience.}
\label{fig:overview}
\end{figure}

\subsection{Overview}
\label{sec:keel-overview}

KEEL is an evidence-grounded framework that converts task experience into
reusable knowledge for subsequent tasks. For task \(P_{i,j}\), the Actor
receives task-conditioned knowledge from \(\mathcal{K}_i^{(j-1)}\) and performs
a single rollout. The verifier evaluates the resulting artifact, and the
Analyzer connects check outcomes to behavior observed in the trajectory.
Supported behavior is distilled into reusable skills, while failures are
converted into corrective lessons. Deterministic managers validate these
updates before adding them to the knowledge banks, ensuring that persistent
knowledge is grounded in both trajectory and verifier evidence. The updated
knowledge is made available only to subsequent tasks.
This design separates environment interaction, semantic knowledge construction,
and deterministic state management.  \Cref{fig:overview} illustrates KEEL's
task-level learning loop.

\subsection{Knowledge Management}
\label{sec:knowledge-management}

\begin{figure}[!t]
\centering
\includegraphics[width=1.0\textwidth]{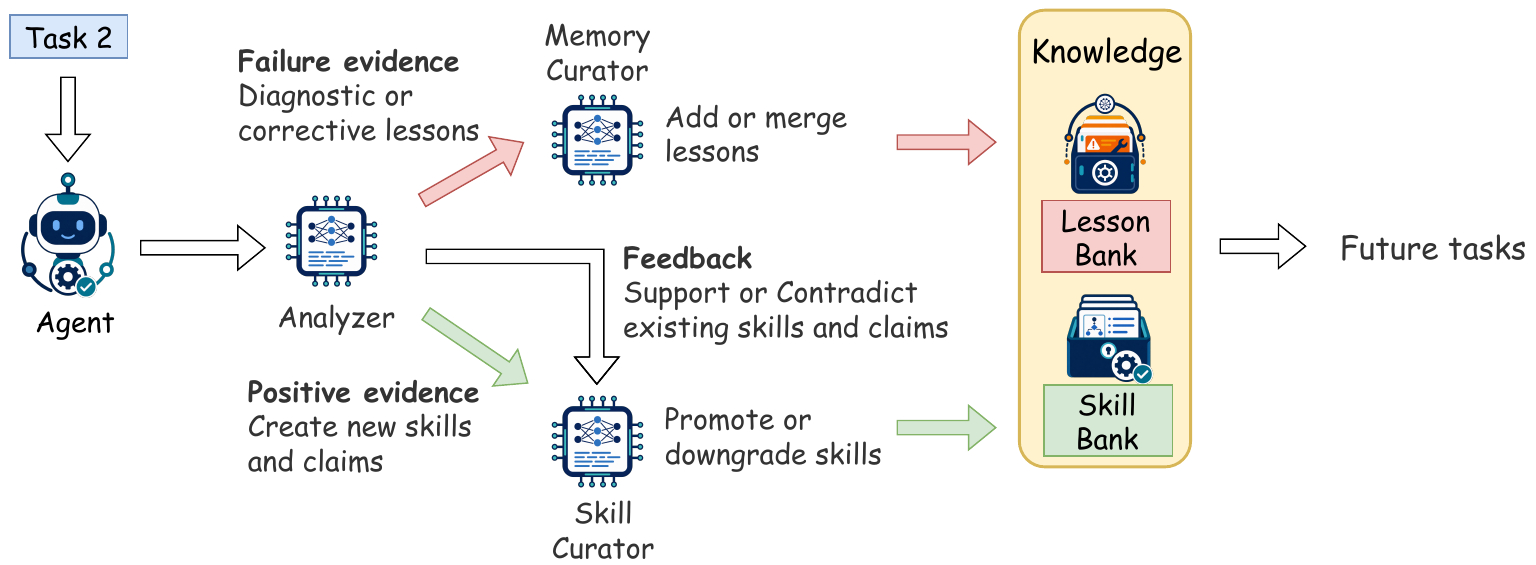}
\caption{Knowledge management in KEEL. The Analyzer separates task experience
into positive evidence, failure evidence, and feedback on existing claims.
The Curators convert positive and failure evidence into candidate skills and
lessons, respectively, while deterministic managers validate all updates before
modifying the persistent knowledge banks.}
\label{fig:zoom}
\end{figure}

\Cref{fig:zoom} zooms in on KEEL's knowledge management process. After each task, KEEL transforms the execution trajectory and verifier outcomes
into knowledge for subsequent tasks. Positive evidence may produce
reusable skills, failure evidence into diagnostic or corrective lessons, and
claim-level feedback informs updates to existing skills.
This decomposition focuses on actionable knowledge rather than restricting the framework to specific information types. For instance, factual constraints and environment conditions may be represented through skill applicability and limitations, while causal hypotheses and exceptions may inform grounded corrective lessons.

\subsubsection{Task Execution and Evidence Attribution}
\label{sec:execution-analysis}

Before executing task \(P_{i,j}\), KEEL extracts requirement-bearing lines
(e.g., output paths, schemas, structural constraints, and preservation rules)
from \(I_{i,j}\) into a requirement ledger
\(L_{i,j}=\operatorname{Ledger}(I_{i,j})\), along with knowledge from the previous family state:

\[
\left(
\widetilde{\mathcal{S}}_{i,j},
\widetilde{\mathcal{M}}_{i,j}
\right)
=
\operatorname{Retrieve}
\left(I_{i,j},\mathcal{K}_i^{(j-1)}\right).
\]
The Actor can access all active, non-suppressed skills, while the deterministic
router may recommend one skill for initial inspection. The lesson view contains
a small subset selected for the current task. Both skills and lessons are
advisory and cannot override the task instruction.

The Actor then performs one rollout:
\[
(\tau_{i,j},\hat{o}_{i,j})
=
\pi_{\theta}
\left(
I_{i,j},
L_{i,j},
\widetilde{\mathcal{S}}_{i,j},
\widetilde{\mathcal{M}}_{i,j}
\right),
\]
where \(\tau_{i,j}\) is the execution trajectory, \(\hat{o}_{i,j}\) is the
output artifact, and the model parameters \(\theta\) remain fixed.


The environment evaluates \(\hat{o}_{i,j}\) and returns check-level verifier
feedback. The Analyzer receives the task instruction, requirement ledger,
execution trajectory, artifact manifest, verifier feedback, current family
repair memory, and the active claims from skills inspected by the Actor. Let
\(X_{i,j}\) denote this input package. The Analyzer produces
\[
\left(
E_{i,j}^{+},
E_{i,j}^{-},
E_{i,j}^{?},
F_{i,j}^{S}
\right)
=
\operatorname{Analyzer}_{\phi}(X_{i,j}).
\]
Here, \(E_{i,j}^{+}\) contains evidence from passed checks,
\(E_{i,j}^{-}\) contains evidence from failed checks, and
\(E_{i,j}^{?}\) contains unresolved or infrastructure-affected evidence.
The feedback \(F_{i,j}^{S}\) classifies inspected skill claims as supported,
contradicted, or not evaluable.

The Analyzer produces one evidence item per verifier check and rates its confidence for each evidence item. Passed checks support
only the tested property, while failed checks do not necessarily reveal a cause
or repair. Invalid Analyzer output and unresolved or infrastructure-affected
evidence cannot update the knowledge banks.

\subsubsection{Skill Curation}
\label{sec:skill-construction}

The Skill Curator converts eligible positive evidence into candidate skills:
\[
\Delta\mathcal{S}_{i,j}
=
\operatorname{SkillCurator}_{\psi}
\left(
E_{i,j}^{+},
\mathcal{S}_i^{(j-1)}
\right).
\]
Each candidate defines its applicability, procedure, verification steps, and
limitations. Skill claims cite supporting checks, evidence references, and
trajectory actions, while procedure and verification steps cite these grounded
claims. Only medium- or high-confidence positive evidence is eligible. The
Curator proposes candidates, and the deterministic Skill Manager decides
whether to accept them.

\subsubsection{Lesson Curation and Memory Management}
\label{sec:lesson-management}

The Memory Curator converts failure evidence and claim-level feedback into up to
two candidate lessons:
\[
\Delta\mathcal{M}_{i,j}
=
\operatorname{MemoryCurator}_{\omega}
\left(
E_{i,j}^{-},
F_{i,j}^{S},
\mathcal{M}_i^{(j-1)}
\right).
\]
A diagnostic lesson records an observed failure, what to inspect, and how to
verify it. A corrective lesson additionally states what to avoid and what to do,
and requires direct causal evidence with medium or high confidence and a
grounded repair hypothesis. Both types cite supporting failed checks and
evidence references. The Memory Manager validates the candidates, merges exact
duplicates, and adds distinct lessons.

Before a later task, KEEL ranks lessons by trigger overlap with the task
instruction, followed by the number of supporting tasks and recency. If no
trigger matches, it selects a fallback using support and recency. 
Unlike skill claims, lessons record context-specific failure observations and corrective guidance, which may remain useful even when subsequent tasks succeed. Therefore, KEEL manages lessons through evidence-based validation, deduplication, and task-conditioned retrieval rather than explicit claim-level state transitions.


\subsection{Skill-Claim Lifecycle}
\label{sec:skill-lifecycle}
\vspace{-0.3em}

\begin{figure}[!t]
\centering
\includegraphics[width=0.8\textwidth]{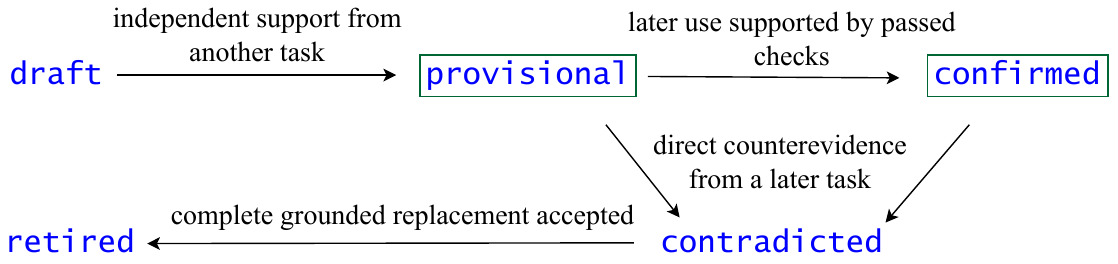}
\caption{Lifecycle of a skill claim. Cross-task evidence promotes a claim,
direct counterevidence quarantines it, and a grounded replacement retires the
old claim while introducing a corrected claim under a new identifier.}
\label{fig:lifecycle}
\end{figure}

A claim about a computational procedure enters as \texttt{draft} after partial success or without
direct evidence of the procedure; otherwise, it enters as
\texttt{provisional}. 
All other accepted claims enter as
\texttt{provisional}.
Draft claims await further support, while provisional claims
are visible to the Actor. Confirmed claims have been validated through later
use. Contradicted and retired claims are hidden but retained for provenance.

As shown in \Cref{fig:lifecycle}, a \texttt{draft} claim becomes \texttt{provisional} when it receives independent support from another task's execution. A \texttt{provisional} claim
becomes \texttt{confirmed} when its later use is supported by relevant passed checks.
Direct medium- or high-confidence counterevidence moves a provisional or
confirmed claim to \texttt{contradicted}, while
\texttt{not\_evaluable} feedback causes no transition.

When an accepted candidate refreshes an existing skill, returned claims preserve
their history, while omitted claims are retired. Contradicted and retired
identifiers cannot be reused. A corrected claim must use a new identifier and
enters as draft or provisional according to its evidence. Before each task, the
Actor receives only provisional and confirmed claims not suppressed by the
selected lesson.

\section{Experimental Results}

\subsection{Experimental Setup}

\vspace{-0.3em}
We evaluate KEEL on the SkillFlow benchmark~\citep{zhang2026skillflow}, using
Terminus-2~\citep{merrill2026terminalbenchbenchmarkingagentshard} (the default implementation included with the SkillFlow codebase \citep{zhangzi-a_skillflow_2026}) as the
execution agent. SkillFlow contains 166 tasks organized into 20 workflow
families. Each family comprises an ordered sequence of related, increasingly
difficult tasks within the same repository.
Tasks are executed in the benchmark-provided order, with one rollout per task.
For all evaluated methods, knowledge is curated independently for each family and is available only to
subsequent tasks in that family.
The benchmark spans diverse domains, including data
science, finance, and life sciences.

We compare three conditions:
\emph{No Skills}, which retains no cross-task knowledge; the skill-curation mechanism released in \emph{SkillFlow}~\citep{zhang2026skillflow}, as a baseline; and \emph{KEEL}.
%
For the direct three-way comparison, we evaluate GPT-5.4, GPT-5.6 Sol, and
Claude Opus 4.6. We additionally evaluate Claude Sonnet 4.6 and Haiku 4.5 to
study the effect of model capability. We use \textit{high} reasoning effort for
GPT-5.6 Sol, \textit{medium} for GPT-5.4, Claude Opus 4.6, and Claude Sonnet
4.6, and disable extended thinking for Claude Haiku
4.5.

\begin{table}[!ht]
\centering
\caption{Task pass rates for Terminus-2 across three models. Parentheses report absolute percentage point changes relative to \textit{No Skills}.}
\label{tab:main-results}
\begin{tabular}{@{}llrrr@{}}
\toprule
Agent & Model & No Skills & SkillFlow & KEEL \\
\midrule
Terminus-2 & Claude Opus 4.6 & 44.0\% & 53.6\% (\textcolor{green!50!black}{\(\uparrow\) 9.6pp}) & \textbf{65.7\% (\textcolor{green!50!black}{\(\uparrow\) 21.7pp})} \\
Terminus-2 & GPT-5.6 Sol & 59.6\% & 63.3\% (\textcolor{green!50!black}{\(\uparrow\) 3.6pp}) & \textbf{65.7\%\ \ \  (\textcolor{green!50!black}{\(\uparrow\) 6.0pp})} \\
Terminus-2 & GPT-5.4 & 39.8\% & 42.2\% (\textcolor{green!50!black}{\(\uparrow\) 2.4pp}) & \textbf{44.0\%\ \ \  (\textcolor{green!50!black}{\(\uparrow\) 4.2pp})} \\
\bottomrule
\end{tabular}
\end{table}



\begin{figure}[!ht]
    \centering
    \begin{subfigure}[t]{0.48\linewidth}
        \centering
        \includegraphics[width=\linewidth]{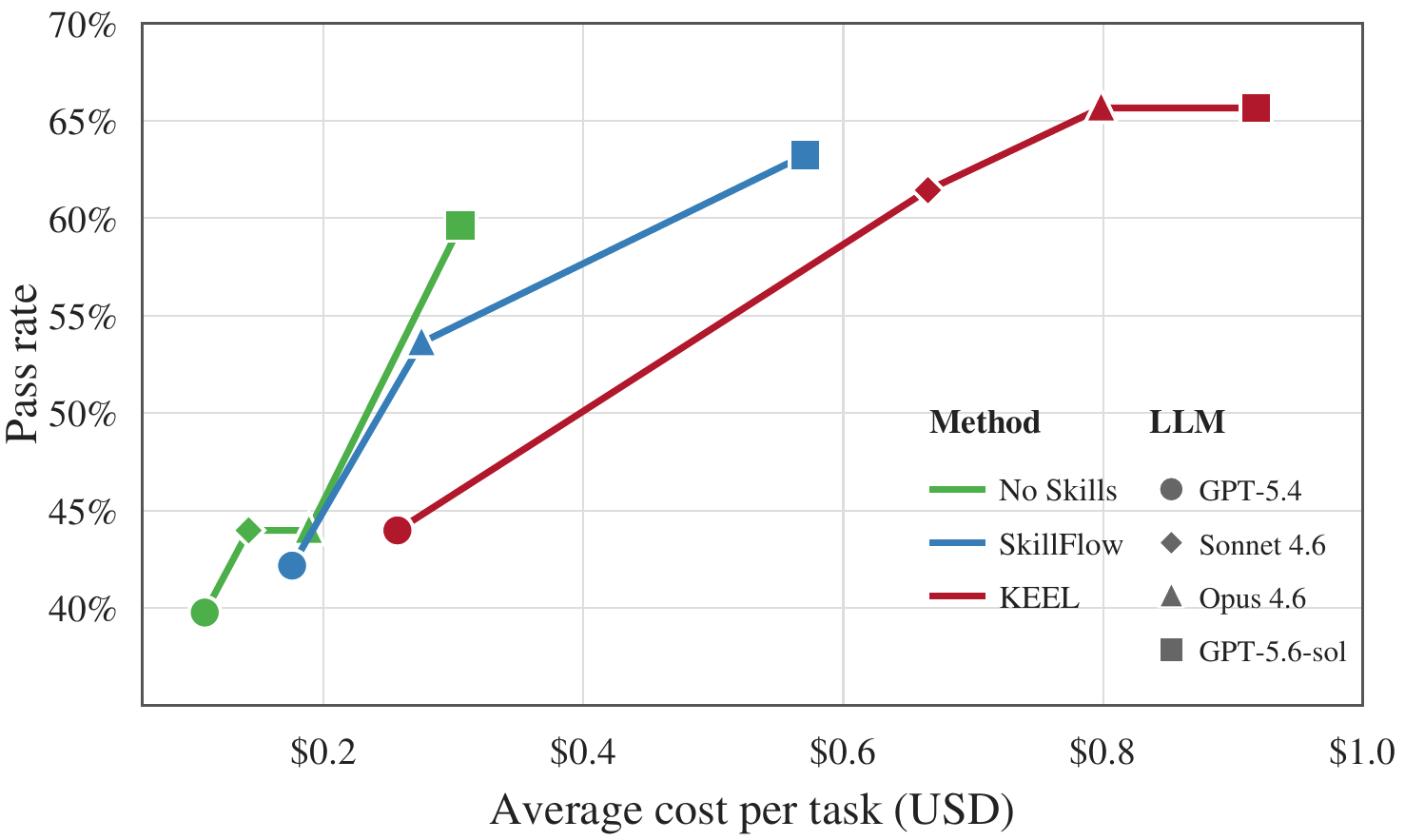}
        \caption{Pass rate versus average inference cost per task.}
        \label{fig:cost-performance}
    \end{subfigure}
    \hfill
    \begin{subfigure}[t]{0.48\linewidth}
        \centering
        \includegraphics[width=\linewidth]{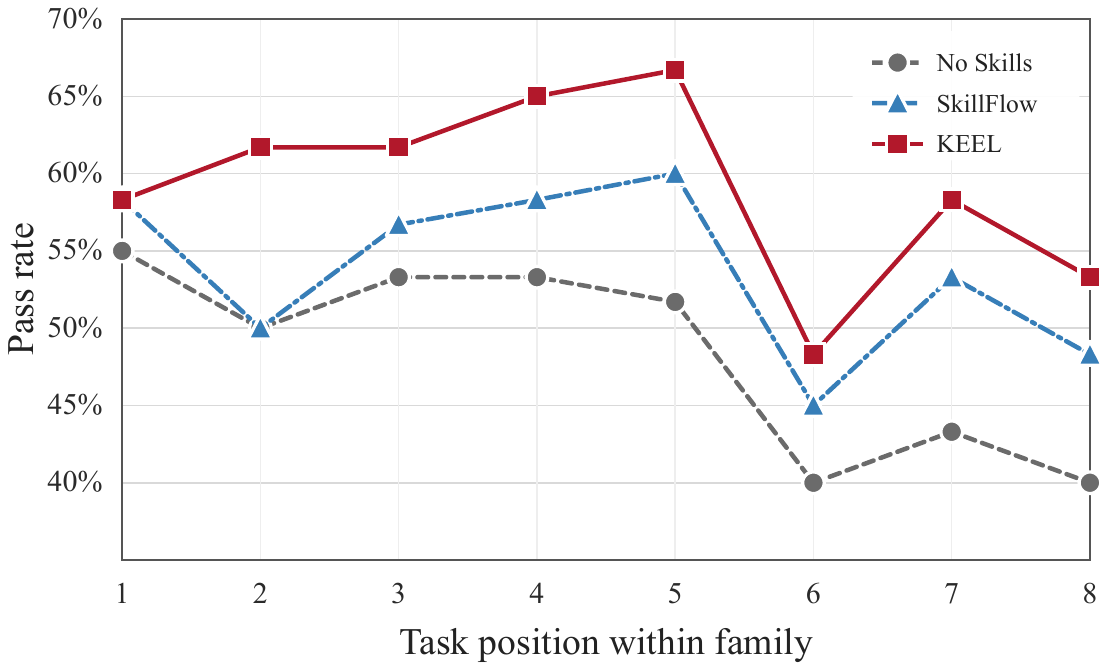}
        \caption{Pass rate by task position within a workflow family.}
        \label{fig:position-pass-rate}
    \end{subfigure}
    \caption{(a) Pass rate versus average inference cost per task. Colors identify methods, and markers identify models. Lines connect results obtained with the same method. (b) Pass rate by task position within a workflow family, aggregated
    across the 20 families and the three models. Positions beyond eight are omitted because they are not shared by all families. }
    \label{fig:combined}
\end{figure}


\subsection{Overall Performance and Cost}

\paragraph{KEEL substantially improves on baseline performance.}
As shown in \Cref{tab:main-results}, across the three models with results for all methods, KEEL improves pass rates
by 4.2 to 21.7 percentage points, compared with 2.4 to 9.6 points for
SkillFlow. The largest gain occurs with Opus 4.6, where KEEL
increases the pass rate from 44.0\% to 65.7\% (a 21.7-point gain), compared
with 53.6\% for SkillFlow (a 9.6-point gain). Its gains are narrower with
GPT-5.6 Sol (6.0 points) and GPT-5.4 (4.2 points),
indicating that the benefit of cross-task knowledge depends on the underlying
model; we investigate this effect further in \Cref{sec:model-ablation}. Thus,
across the models evaluated, more extensive knowledge management consistently
translates into higher task success, although the magnitude of the improvement
varies substantially by model.


\vspace{-0.2em}

\paragraph{KEEL's performance gains come with additional inference cost.}
\Cref{fig:cost-performance} shows that KEEL achieves the highest pass rate for each model, but requires additional inference to construct, validate, and apply its richer knowledge state. For example, on Opus 4.6, KEEL improves pass rate by about 50\% (relative) over No Skills while increasing estimated cost per instance from \$0.19 to \$0.80. The additional cost stems primarily from increased token usage, which is inherent to KEEL's current design, and secondarily from reduced cache efficiency, which could likely be improved beyond our prototype implementation. KEEL therefore provides a mechanism for trading additional inference-time computation for substantially stronger continual-learning performance.

\subsection{Task-Level Evidence of Cross-Task Learning}
\label{sec:task-level-analysis}

Aggregate pass rates do not reveal which individual task outcomes change.
We therefore examine performance on individual tasks as
knowledge accumulates within a family.





\vspace{-0.5em}
\paragraph{KEEL's advantage emerges after cross-task knowledge becomes available.}
As shown in \Cref{fig:position-pass-rate}, at the first task position -- before either knowledge-based method has learned
from an earlier task -- KEEL and SkillFlow obtain the same aggregate pass rate.
From the second position onward, KEEL outperforms both baselines at every
position. Averaged over positions 2-8, KEEL exceeds SkillFlow's pass rate by 6.2 percentage points and
No Skills by 11.9 points. This pattern is consistent with KEEL benefiting from
knowledge accumulated across earlier tasks in the same family, rather than
providing an advantage independent of task history.


\FloatBarrier


\subsection{Model and Reasoning Configuration}
\label{sec:model-ablation}
\paragraph{KEEL's benefits depend on the capability of the underlying model.}
From \Cref{tab:model-strength}, we observe that KEEL improves Opus 4.6 from 44.0\% to
65.7\% and Sonnet 4.6 from 44.0\% to 61.4\%, but decreases
Haiku 4.5 from 37.3\% to 28.9\%. This suggests that KEEL's
richer knowledge can provide substantial leverage to sufficiently capable
models, while weaker models may be less able to reliably interpret and apply
the additional context. This dependence may arise during knowledge
construction as well as execution. KEEL's Analyzer relates check-level
verifier feedback to evidence in the execution trajectory before skills and
lessons are constructed or revised. Because its outputs directly inform these
updates, reliable Analyzer output is a prerequisite for reliable knowledge
evolution. KEEL may therefore benefit from using a sufficiently capable LLM or
a stronger reasoning configuration for evidence analysis.

\begin{table}[!t]
\begin{minipage}{.55\linewidth}
\centering
\caption{Task pass rates across Claude model and reasoning configurations.
Opus 4.6 and Sonnet 4.6 use medium reasoning, whereas Haiku 4.5 has extended
thinking disabled. Parentheses report changes relative to No Skills.}
\label{tab:model-strength}
\begin{tabular}{@{}lrr@{}}
\toprule
Model & No Skills & KEEL \\
\midrule
Claude Opus 4.6 & 44.0\% & 65.7\% (\textcolor{green!50!black}{\(\uparrow\) 21.7pp}) \\
Claude Sonnet 4.6 & 44.0\% & 61.4\% (\textcolor{green!50!black}{\(\uparrow\) 17.5pp}) \\
Claude Haiku 4.5 & 37.3\% & 28.9\% (\textcolor{red!70!black}{\(\downarrow\) 8.4pp}) \\
\bottomrule
\end{tabular}
\end{minipage}\quad%
\begin{minipage}{.45\linewidth}
\centering
\caption{Ablation of KEEL's skills/memories using
Terminus-2 with Claude Opus 4.6, compared with baselines.}
\label{tab:keel-ablation}
\begin{tabular}{@{}lr@{}}
\toprule
Configuration & Pass rate \\
\midrule
Full KEEL & 65.7\% \\
KEEL-MemoryOnly & 63.9\% (\textcolor{red!70!black}{\(\downarrow\) 1.8pp}) \\
KEEL-SkillsOnly & 59.0\% (\textcolor{red!70!black}{\(\downarrow\) 6.6pp}) \\
\midrule
SkillFlow & 53.6\% (\textcolor{red!70!black}{\(\downarrow\) 12.1pp}) \\
No skills & 44.0\% (\textcolor{red!70!black}{\(\downarrow\) 21.7pp}) \\
\bottomrule
\end{tabular}
\end{minipage}
\end{table}

\FloatBarrier

\subsection{Ablation}
\label{sec:component-ablation}
\paragraph{Skills and lessons provide complementary benefits.}
Full KEEL achieves the highest pass rate at 65.7\%. \Cref{tab:keel-ablation} shows that removing skills reduces it
to 63.9\%, while removing lessons reduces it to 59.0\%, indicating that both
components contribute. The larger decrease without lessons suggests that
failure-derived guidance provides the greater marginal benefit in this Opus
4.6 experiment.  KEEL-SkillsOnly outperforms baseline SkillFlow skill curation by 5.4 points, 
suggesting that KEEL's evidence-grounded skill-management pipeline is beneficial.

\section{Conclusion}
\vspace{-0.5em}
We introduced KEEL, a continual-learning framework that treats external agent
knowledge as an evolving state rather than an append-only memory. KEEL grounds
verifier outcomes in execution evidence, converts supported behavior into
reusable skills, and records failures as diagnostic or corrective lessons.
Individual knowledge claims are subsequently confirmed, contradicted, or
retired as evidence accumulates.

We evaluated KEEL on the SkillFlow benchmark, which contains 166 tasks
organized into 20 ordered workflow families. KEEL outperforms both the No
Skills baseline and the original skill-curation method proposed in the
SkillFlow paper across all complete model comparisons. Further analysis shows that skills and lessons
provide complementary benefits, and that KEEL's advantage emerges after a single task is completed.

The limitation is that knowledge quality depends on reliable evidence
attribution by the Analyzer. Future work should decompose its responsibilities
into separately evaluated components, such as evidence extraction, action
attribution, and causal diagnosis, and investigate whether a stronger
Analyzer can support a less expensive execution model.

\clearpage
\newpage

\bibliographystyle{wenzhe_num}
\bibliography{reference}

\clearpage
\newpage

\newpage

\end{document}